\documentclass[lettersize,journal]{IEEEtran}
\usepackage{amsmath,amsfonts}
\usepackage{algorithmic}
\usepackage{algorithm}
\usepackage{array}
\usepackage[caption=false,font=normalsize,labelfont=sf,textfont=sf]{subfig}
\usepackage{textcomp}
\usepackage{stfloats}
\usepackage{url}
\usepackage{verbatim}
\usepackage{graphicx}
\usepackage{cite}
\usepackage{booktabs}
\usepackage{multirow}
\usepackage{comment}
\usepackage{makecell}
\usepackage{cuted}
\usepackage{capt-of}

\usepackage{pifont}% http://ctan.org/pkg/pifont

\usepackage{url}
\usepackage[table]{xcolor}
\definecolor{graybg}{gray}{0.93}

\begin{document}

\title{No-Reference Rendered Video Quality Assessment: Dataset and Metrics}

\author{
Sipeng Yang,
Jiayu Ji,
Qingchuan Zhu,
Zhiyao Yang,
% Kede Ma, 
Xiaogang Jin,~\IEEEmembership{Member,~IEEE}% <-this % stops a space
\IEEEcompsocitemizethanks{\IEEEcompsocthanksitem Sipeng Yang, Jiayu Ji, Qingchuan Zhu, and Xiaogang Jin are with the State Key Lab of CAD\&CG, Zhejiang University, Hangzhou, P. R. China.
\IEEEcompsocthanksitem Zhiyao Yang is with OPPO Nanjing Research Center, Nanjing, P. R. China.
% \IEEEcompsocthanksitem Kede Ma is with the City University of Hong Kong, Hong Kong.
\IEEEcompsocthanksitem Xiaogang Jin is the corresponding author. E-mail: jin@cad.zju.edu.cn}% <-this % stops an unwanted space
\thanks{Manuscript received MM DD, 2024; revised MM DD, 2024.}}

\markboth{Journal of \LaTeX\ Class Files,~Vol.~14, No.~8, August~2021}%
{Shell \MakeLowercase{\textit{et al.}}: A Sample Article Using IEEEtran.cls for IEEE Journals}


\maketitle

\begin{abstract}
Quality assessment of videos is crucial for many computer graphics applications, including video games, virtual reality, and augmented reality, where visual performance has a significant impact on user experience. When test videos cannot be perfectly aligned with references or when references are unavailable, the significance of no-reference video quality assessment (NR-VQA) methods is undeniable. However, existing NR-VQA datasets and metrics are primarily focused on camera-captured videos; applying them directly to rendered videos would result in biased predictions, as rendered videos are more prone to temporal artifacts. To address this, we present a large rendering-oriented video dataset with subjective quality annotations, as well as a designed NR-VQA metric specific to rendered videos. The proposed dataset includes a wide range of 3D scenes and rendering settings, with quality scores annotated for various display types to better reflect real-world application scenarios. Building on this dataset, we calibrate our NR-VQA metric to assess rendered video quality by looking at both image quality and temporal stability. We compare our metric to existing NR-VQA metrics, demonstrating its superior performance on rendered videos. Finally, we demonstrate that our metric can be used to benchmark supersampling methods and assess frame generation strategies in real-time rendering. 
\end{abstract}


\section{Introduction}
\IEEEPARstart{V}{ideo} quality metrics are essential for optimizing rendering pipelines to ensure high-fidelity outcomes in rendered content \cite{flip}. Well-known metrics, such as structural similarity (SSIM)~\cite{wang2004image} and peak signal-to-noise ratio (PSNR), along with human visual system (HVS)-based methods~\cite{flip, mantiuk2021fovvideovdp}, require perfectly aligned and high-quality reference images to evaluate the similarity between test images and references. 
However, in commercial graphics rendering applications, misalignment between test and reference images frequently occurs due to the difficulty in consistently replicating exact camera positions and dynamic object poses across different rendering cycles, which reduces the effectiveness of full-reference metrics. These issues highlight the critical need for reliable no-reference video quality assessment (NR-VQA) metrics.

% Due to random movements within packaged game assets, difficulty in accurate camera trajectory control in packaged rendering programs, and discrepancies between motion estimation outputs and the ground truth, all of which reduce the effectiveness of full-reference metrics. 

\begin{figure}
\begin{center}
\includegraphics[width=0.96\linewidth]{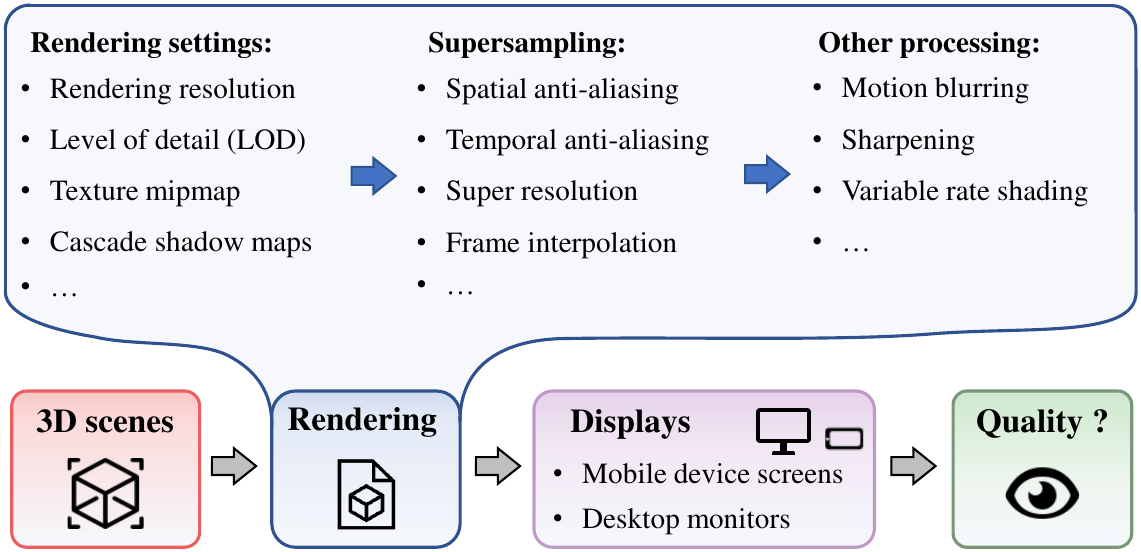}
\end{center}
\caption{
Process for assessing the quality of rendered videos. Rendered videos undergo a complex transition from 3D scenes to on-screen displays. During this transition, 3D scene quality, rendering configurations, and display types all have a significant impact on perceived quality. 
% Process of rendered video quality assessment. Rendered videos undergo a complex transition from 3D scenes to on-screen displays. During this transition, the 3D scene quality, rendering configurations, and display types all significantly influence the perceived quality.
}
\label{fig:teaser}
\end{figure}

NR-VQA methods aim to evaluate the perceptual quality of test videos without relying on references. Extensive research has been conducted in this field, with datasets such as KoNViD-1k~\cite{kon-1kdataset} and LIVE-VQC~\cite{dataset-LIVEVQC}, as well as developed NR-VQA metrics focusing on temporal aggregation~\cite{GSTVQA21, richdatabase2021}, and temporal pooling~\cite{sun2022deep, wu2023exploring}. These efforts primarily target the quality assessment of camera-captured videos, which are usually degraded by \textit{blurring}, \textit{noise}, and \textit{overexposure} issues~\cite{vsfa19}. However, the main factors affecting the perceptual quality of rendered videos differ from those impacting camera-captured ones. Rendered videos, limited by per-pixel sampling rates, are particularly susceptible to temporal artifacts like \textit{flickering} and \textit{moving jaggies}. These artifacts are especially pronounced and disruptive due to the human visual system's acute sensitivity to temporal variations~\cite{biological-cells}. As a result, there is an urgent need to create a new dataset and a specifically designed metric for NR-VQA of rendered videos.%, which includes a comprehensive analysis of rendering and display processes.

Developing NR-VQA methods for rendered videos first requires addressing the scarcity of suitable datasets. The perceived quality of rendered videos is influenced by multiple factors, including 3D scene resources, rendering pipelines, and display devices, as shown in Fig.~\ref{fig:teaser}. Ensuring these aspects are represented in the dataset is crucial to effectively reflect real-world conditions.
Besides the dataset, designing effective NR-VQA metrics for rendered videos poses additional challenges. For rendered content, recent convolutional neural network (CNN)-based models ~\cite{cardoso2022training} have leveraged extra G-buffers to detect \textit{static artifacts}, such as aliasing and Moir\'{e} effects. However, detecting \textit{temporal artifacts} or assessing temporal stability in dynamic video scenes, especially in the absence of reference images, remains a significant unresolved challenge. As illustrated in Fig.~\ref{fig:res1}, failing to consider the temporal stability of videos can result in significant biases in NR-VQA of rendered videos.

To address the challenges, we introduce a large rendering-oriented dataset accompanied by a new NR-VQA metric tailored for rendered videos. The dataset, named \textit{ReVQ-2k}, features a diverse array of 3D scenes and rendering configurations, with perceptual quality scores annotated on both smartphone screens and desktop displays. To enhance the evaluation on video temporal stability, we expand our collection of subjective annotations to include not only overall quality scores but also temporal stability scores, providing additional supervision for metric calibration. Simultaneously, we propose an NR-VQA metric that evaluates rendered video quality from two perspectives. First, we focus on image quality factors such as clarity and static artifacts, aligning with existing NR-VQA methods \cite{wu2023neighbourhood, wu2023exploring}. Given the extensive analysis in existing literature, we directly adopt a state-of-the-art (SOTA) practice, FAST-VQA~\cite{wu2023neighbourhood}, for the assessment of the image quality aspect. Second, for the evaluation of temporal artifacts, we propose utilizing motion estimation to counteract object movement across frames, and then using a multi-timescale image differencing module to assess the temporal stability of videos. This module can be calibrated using the annotated temporal stability scores from ReVQ-2k to achieve better precision. Finally, our metric integrates these two aspects to provide a comprehensive evaluation of rendered video quality, offering more accurate predictions than existing metrics.

We demonstrate the utility of our calibrated NR-VQA metric through practical applications, including comparing the video quality of closed-source supersampling methods and assessing the perceived quality of frame generation strategies. Our metric provides stable quantitative assessments, offering substantial advantages over manual annotations for many relevant real-world applications.

The main contributions of the paper are as follows:
\begin{itemize}% [leftmargin=21pt] 
    \item We introduce a large rendering-oriented dataset, \textit{ReVQ-2k}, comprising 2,000 rendered videos that feature a wide range of scenes and rendering settings, along with 57,450 subjective annotations for overall quality and temporal stability on various displays.
    \item We develop a novel NR-VQA metric that considers both image quality and temporal stability of rendered videos.
    \item  The utility of the proposed metric is demonstrated in real-world tasks, including the evaluation of closed-source supersampling methods and frame generation strategies.
\end{itemize}

% Extensive experiments demonstrate that our proposed method surpasses other state-of-the-art (SOTA) approaches on both the new rendered dataset and existing realistic VQA datasets.
% \textbf{Open-source dataset for NR-VQA of rendered videos.} 
% \textbf{Novel NR-VQA metric for rendered videos.}
% \textbf{Real-world applications validation.}

\section{Related Works}
\subsection{NR-VQA Datasets}
High-quality datasets are essential for the calibration and evaluation of NR-VQA metrics. Early datasets \cite{syn-dataset1, syn-dataset2} in this field primarily focus on videos affected by specific distortions from compression and transmission processes. These datasets are limited to a narrow range of distortions, which restricts their utility in real-world scenarios \cite{patchvq}. In contrast, subsequent studies like CVD2014 \cite{dataset-cvd2014}, LIVE-Qualcomm \cite{dataset-liveQualcomm}, and KoNViD-1k \cite{kon-1kdataset} have collected a wide variety of videos and provided manual quality ratings, significantly enhancing the diversity and practical relevance of these resources. These datasets have become instrumental in advancing NR-VQA research, particularly for assessing real-world video quality. However, rendered videos, which are increasingly prevalent in various applications, are scarcely represented within these datasets.

Despite a recent work, LIVE-YT-Gaming \cite{yu2023subjective}, introduces a dataset for video streams VQA of gaming content, it primarily evaluates the impact of video compression, without considering rendering settings and display impacts as illustrated in Fig. \ref{fig:teaser}. Our work aims to fill this gap by collecting data with a broad range of 3D scenes and rendering settings, along with providing quality ratings for videos displayed on various types of screens. This initiative significantly enhances the capability of NR-VQA metrics to assess the quality of rendered videos, providing crucial resources for quality evaluation in this field.

% , and there is no specific dataset for manually rating rendered video quality

\begin{table*}
\centering
\renewcommand{\arraystretch}{1.3}
\caption{Summary of existing NR-VQA datasets and the proposed ReVQ-2k dataset.}
\footnotesize
\begin{tabular}{c|c|c|c|c|p{7.6cm}} 
\hline \hline
\textbf{Dataset} & \textbf{Scale} & \textbf{Duration} & \textbf{Resolution} & \textbf{Frame Rate} & \textbf{Content Origin} \\
\hline
CVD2014 \cite{dataset-cvd2014}       		& 234   & 10-25 sec & 480p, 720p		& 10-32         & Captured with real cameras, no post-processing distortions. \\
LIVE-Qualcomm~\cite{dataset-liveQualcomm}	& 208  	& 15 sec    & 1080p   			& 30          	& Captured with mobile devices.\\
KoNViD-1k \cite{kon-1kdataset}          	& 1,200  & 8 sec     & 540p   			& 24, 25, 30	& Camera-captured in-the-wild videos.\\
LIVE-VQC \cite{dataset-LIVEVQC}      		& 585   & 10 sec    & 240p-1080p   		& 20-30        	& Captured using 101 different camera devices.\\
LIVE-YT-Gaming~\cite{yu2023subjective}  	& 600   & 8-9 sec  	& 360p-1080p		& 30, 60        & Compressed streaming game videos.\\\hline
\rowcolor{graybg}\textbf{ReVQ-2k (ours)}             & 2,000  & 8 sec     & 720p, 1080p, 2k	& 60            & Rendered videos using various rendering settings, evaluated on different types of displays.\\
\hline \hline
\end{tabular}
\label{table:datasets}
\end{table*}

\subsection{NR-VQA metrics}
% Existing video quality assessment (VQA) methods are typically categorized into three types: full-reference VQA~\cite{wang2004video}, reduced-reference VQA~\cite{soundararajan2012video}, and NR-VQA \cite{ugc-vqa}. As aligned reference videos are usually unavailable or nonexistent in practical tasks, NR-VQA has received considerable interest due to its practical utility \cite{GSTVQA21}. In the following, we mainly discuss key NR-VQA methods related to our approach.

\textbf{Classical NR-VQA Methods.}
Over the past decades, extensive efforts in NR-VQA have been conducted from diverse perspectives. A prominent strategy involves quantifying artifacts in test videos. Early studies thoroughly examine the impact of image artifacts such as blocking \cite{blocking}, noise \cite{noise1}, and blurring \cite{blurring} on perceived video quality. Additionally, alternative approaches assess video quality using visual indicators like local contrast, brightness, and colorfulness \cite{colorfulness1, colorfulness2}. These NR-VQA approaches leverage hand-crafted features, which enhance the model interpretability.
However, they often yield biased evaluations when confronted with diverse video content and distortion types.

\textbf{Deep Learning-Based NR-VQA Methods.}
To improve evaluation performance across various video categories, recent NR-VQA methods have shifted towards leveraging large video datasets with manually annotated quality labels to train deep learning models for automatic prediction of perceived video quality. Due to the superior feature extraction capabilities of deep neural networks (DNNs), models based on these networks generally outperform those relying on hand-crafted features in terms of accuracy. Among the various techniques employed, the gated recurrent unit (GRU)-based module, capable of learning long-term dependencies in sequential data, is prevalent in NR-VQA algorithms to model the temporal features of videos \cite{vsfa19, mixtrain2021}. Other temporal modeling methods, including pyramid temporal aggregation \cite{GSTVQA21}, motion features statistic \cite{CNN-TLVQM20}, 3D CNNs \cite{zhang2023md}, and SlowFast networks \cite{feichtenhofer2019slowfast}, have also proven effective in NR-VQA for temporal information modeling.

Extensive experiments have demonstrated the effectiveness of DNNs-based NR-VQA metrics on real-world datasets \cite{wu2023exploring}. However, these methods are generally not designed to account for rendering-specific artifacts such as moving jaggies and flickering \cite{renderingflicker, CGF-renderingArtifacts}, which are prevalent in rendered videos. These limitations motivate us to develop a new NR-VQA metric that is specifically designed for predicting rendered video quality.

\subsection{Full-Reference Metrics for Rendered Videos}
In addition to no-reference solutions, full-reference metrics have been specifically designed and widely used to evaluate rendered video quality.
% Beyond no-reference solutions, there are full-reference metrics specifically designed or widely used for evaluating rendered video quality. 
These methods can be broadly categorized into two types: similarity measures and artifact detection-based approaches. 
Similarity measures, such as SSIM \cite{wang2004image}, PSNR, and root mean square error (RMSE), operate on the assumption that greater similarity to a reference image indicates higher content quality, whereas artifact detection-based methods concentrate on the impact of specific artifacts on rendered video quality.
%Similarity measures such as SSIM \cite{wang2004image}, PSNR, and root mean square error (RMSE) operate on the principle that greater similarity to a reference image indicates higher content quality, and are extensively utilized in graphics applications. 
 Due to the scarcity of training data with manual annotations, methods like the contrast sensitivity function (CSF) \cite{barten2003formula} are frequently employed for visual distortion detection. For instance, Aydin \textit{et al.} \cite{aydin2010video} use a 3D CSF and psychometric function metrics to assess distortion visibility in computer-generated videos. Another study \cite{aydin2008dynamic} proposes a calibrated human visual system model for predicting distortion maps in high dynamic range images. Mantiuk \textit{et al.} \cite{mantiuk2021fovvideovdp} have developed a per-pixel visual difference predictor to compare reference and distorted video sequences. More recently, deep learning networks have been employed to detect artifacts in rendered content by training on image datasets with localized distortion maps for accurate visible distortion prediction \cite{wolski2018dataset, cardoso2022training}. However, these methods depend on ground-truth references or rendered G-buffers and fail to assess the temporal stability of videos \cite{yang2020survey}, leaving a gap in research specifically for NR-VQA of rendered videos.

% \begin{figure}
% \centering
% \includegraphics[width=8cm]{pictures/RVQ-400-no-score.pdf}
% \caption{Sample video frames from various scenarios with varying rendering quality settings. From left to right, the rendered quality gradually gets better.} %The rendered quality gradually improves from left to right.}
% \label{fig:RVQ-400}
% \end{figure}

% . To our knowledge, this is the first dataset dedicated to no-reference rendering-oriented video quality assessmen

\begin{figure*}
\centering
\includegraphics[width=0.95\linewidth]{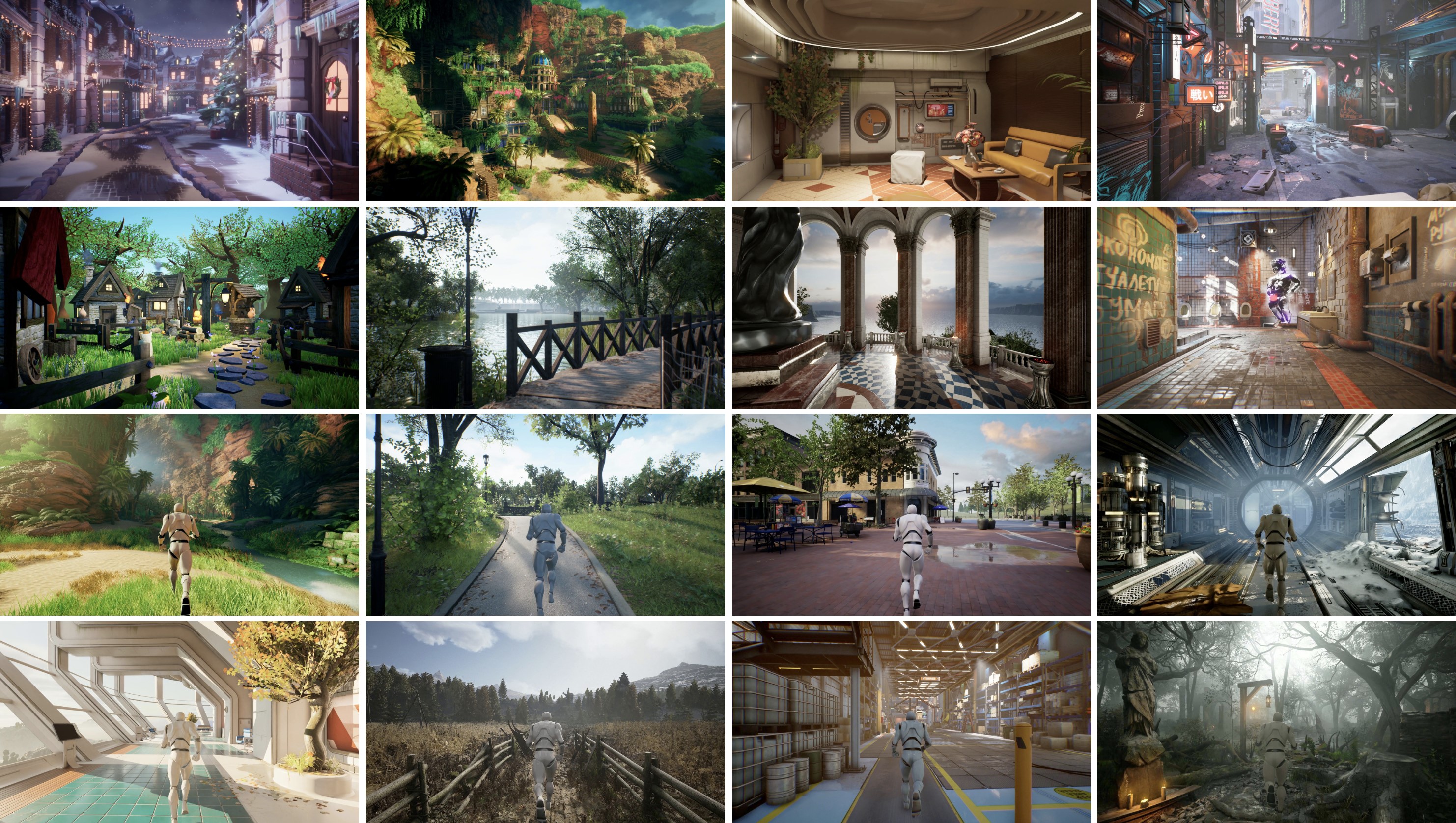}
\caption{Examples of 3D scenes from our ReVQ-2k dataset, featuring urban, interior, and landscape environments under different weather conditions.}
\label{fig:dataset}
\end{figure*}

\begin{figure*}
\centering
\includegraphics[width=\linewidth]{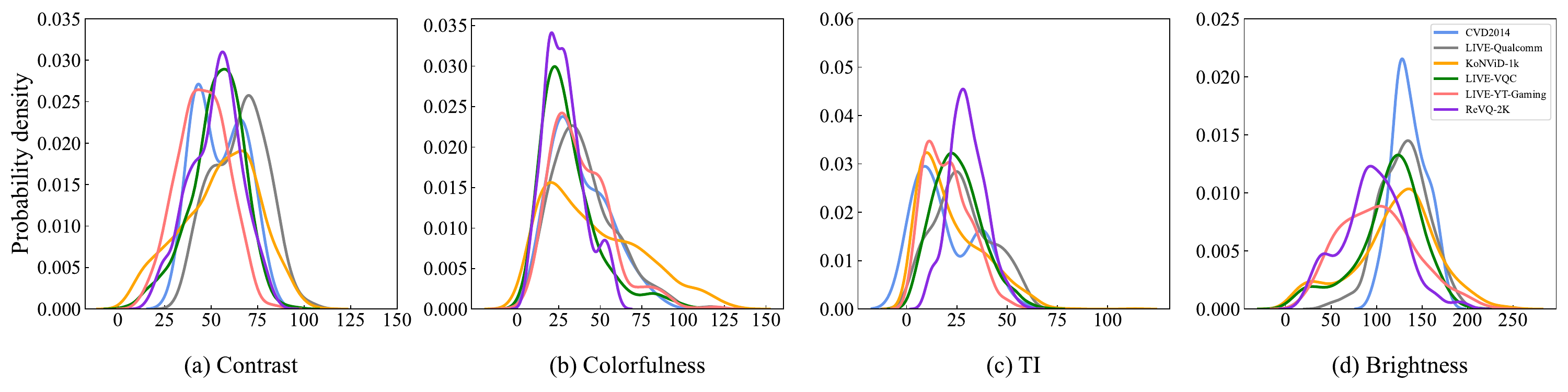}
\caption{Distributions of low-level quantitative attributes of our proposed ReVQ-2k dataset and five existing datasets.}
\label{fig:distri}
\end{figure*}

\section{ReVQ-2k Dataset}
We begin by introducing the proposed rendered video dataset and the subjective quality study conducted on it. Our dataset, known as rendered video quality-2k (ReVQ-2k), includes 2,000 rendered video clips and 57,450 subjective quality annotations. Tab.~\ref{table:datasets} provides a detailed comparison of existing datasets.
% We first introduce the proposed rendered video dataset and the subjective quality study conducted on it. The dataset, termed rendered video quality-2k (ReVQ-2k), comprises 2,000 rendered video clips along with 57,450 subjective quality annotations. Comparative information related to existing datasets is detailed in Tab.~\ref{table:datasets}.

\subsection{Video Collection}
Our data collection strategy is guided by the analysis presented in Fig.~\ref{fig:teaser}, which outlines the origins of rendered videos. This strategy incorporates a wide range of 3D scenes and objects, resulting in a rich diversity of rendering styles and content.
%This strategy includes a broad spectrum of 3D scenes and objects, ensuring a rich diversity in rendering styles and content. 
Moreover, our rendering pipeline utilizes a comprehensive suite of rendering configurations, supplemented by various supersampling and post-processing techniques, to accurately replicate real-world scenarios.

\textbf{3D Scenes.} The videos in our ReVQ-2k dataset are created using Unreal Engine 4 (UE4)~\cite{EU} and Unreal Engine 5 (UE5)~\cite{unrealengine}. We select 15 diverse 3D scenes to encompass a broad range of visual environments. These environments include urban street scenes, interior settings, outdoor landscapes, and scenes rendered in a cartoon style. To further enhance diversity, we capture scenes at various times of the day, such as night and noon, as well as under different weather conditions, such as dusk and snow. Fig.~\ref{fig:dataset} showcases examples of the scenes included in our dataset.

\textbf{Rendering Settings.} Unreal Engine allows for extensive rendering pipeline customization. When creating the ReVQ-2k dataset, we choose settings such as view distances, anti-aliasing methods, post-processing effects, shadows, textures, effect quality, and resolution scaling. These videos are generated at various scalability levels, with each using a random combination of adjustable settings. We also use popular supersampling techniques for video generation, such as FidelityFX super resolution (FSR) \cite{fsr2}, deep learning super sampling (DLSS) \cite{liu2020dlss}, and temporal anti-aliasing upscaling (TAAU) \cite{ue42020taau}. The project homepage includes detailed information about the rendering settings.
% Unreal Engine provides extensive customization capabilities for the rendering pipeline. In creating the ReVQ-2k dataset, we select settings that include view distances, anti-aliasing methods, post-processing effects, shadows, textures, effect quality, and resolution scaling, among others. These videos are generated across various scalability levels, each employing a random combination of adjustable settings. Additionally, we incorporate popular supersampling techniques for video generation, including FidelityFX super resolution (FSR) \cite{fsr2}, deep learning super sampling (DLSS) \cite{liu2020dlss}, and temporal anti-aliasing upscaling (TAAU) \cite{ue42020taau}. Detailed information on the rendering settings is available on the project homepage.

With the rendering settings established, we can now begin collecting rendered videos. We use three resolution settings on the selected 3D scenes: 720p, 1080p, and 2K. For each resolution setting, approximately 700 video clips are gathered. The 720p videos are optimized for smartphone screens, whereas the 1080p and 2K videos are designed for desktop monitors, reflecting typical real-world usage scenarios.
% With the rendering settings established, we now proceed with the collection of rendered videos. From the selected 3D scenes, we employ three resolution settings: 720p, 1080p, and 2K. For each resolution setting, approximately 700 video clips are collected. The 720p videos are specifically prepared for display on smartphone screens, while the 1080p and 2K videos are intended for desktop monitors, reflecting typical real-world usage scenarios.

% Videos produced using these algorithms are included in both the training and test sets of ReVQ-2k. At last, the dataset also includes videos processed with other prevalent rendering or video processing techniques, such as variable rate shading \cite{vrs2021}, sharpening, and video compression \cite{richardson2004h}
% 10 are randomly chosen to serve as the data source for the training set, while the remaining 4 scenes are allocated to the test set. For the training set, approximately 150 videos are collected from each of the 10 scenes, with each video being 8 seconds in duration and recorded at 60 FPS. 

% For the test set, across its 4 scenes, we collect 300 videos of 8 seconds at 1080p resolution for each scene, distributing 100 videos across 720P, 1080P, and 2K resolutions to support test on various display devices.

\textbf{Data Analysis.} To validate the diversity of the ReVQ-2k dataset, we analyze its low-level quantitative attributes \cite{ugc-vqa} and compare them to those of established datasets. Attributes such as contrast, colorfulness, temporal information (TI), and brightness provide a comprehensive comparison of diversity among these datasets. Fig.~\ref{fig:distri} illustrates the distributions of these attributes across various datasets, including ReVQ-2k, CVD2014 \cite{dataset-cvd2014}, KoNViD-1k \cite{kon-1kdataset}, LIVE-Qualcomm \cite{dataset-liveQualcomm}, LIVE-VQC \cite{dataset-LIVEVQC}, and LIVE-YT-Gaming \cite{yu2023subjective}. Our analysis shows that ReVQ-2k features a wide range of contrast and colorfulness, aligning with existing datasets. It also exhibits high levels of TI, indicating more frequent and larger camera movements in ReVQ-2k videos. Although its brightness is slightly lower compared to other datasets, the difference is marginal. Given the rich and varied content of ReVQ-2k, we believe that it is well-suited for calibrating and evaluating VQA metrics, akin to existing datasets.

% , more than half of whom have experience in computer graphics or video processing

\subsection{Subjective Quality Study}
\subsubsection{Quality Scores: OA-MOS and TS-MOS}
To quantify the perceived video quality within the ReVQ-2k dataset, we employ the mean opinion score (MOS) in our user study, which offers an intuitive reflection of user-perceived quality. Our subjective quality assessment utilizes the overall mean opinion score (\textbf{OA-MOS}) for a comprehensive evaluation of rendered video quality. The OA-MOS, widely recognized in prior NR-VQA research \cite{dataset-youtubeUGC, dataset-cvd2014, dataset-liveQualcomm}, reflects the overall quality of videos by considering factors such as colorfulness, block artifacts, visual blurring, and temporal disruptions like flickering. Additionally, to address the specific challenges of assessing temporal stability of  rendered videos, we introduce a new measure, temporal stability mean opinion score (\textbf{TS-MOS}). This measure evaluates the temporal stability of videos, focusing on issues like flickering, moving jaggies, and other temporal artifacts that significantly affect rendered video quality \cite{renderingflicker}. Incorporating TS-MOS into the evaluation adds extra oversight, allowing video quality prediction models to more accurately assess and predict the quality of rendered videos.
% This measure specifically evaluates the temporal stability of videos, focusing on issues like flickering, moving jaggies, and other temporal artifacts that critically affect rendered video quality \cite{renderingflicker}. Incorporating TS-MOS into our evaluation framework provides additional supervision, enabling video quality prediction models to more precisely assess and predict the quality of rendered videos.

\begin{table*}
\centering\renewcommand{\arraystretch}{1.3}
\footnotesize
\caption{Detailed annotation criteria for subjective video quality scoring.}
\label{tab:mos_criteria}

\begin{tabular}{p{1.4cm}|p{6.5cm}|p{8.8cm}}
\hline \hline
\textbf{Score} & \textbf{TS-MOS Criteria} & \textbf{OA-MOS Criteria} \\
\hline
1 Bad & Significant temporal instabilities, flickering, or severe ghosting disrupt content continuity. & Difficulty to follow due to pronounced noise, block artifacts, Moir\'{e} patterns, blurring, substantial flickering, and lag. \\
\hline
2 Poor & Noticeable temporal inconsistencies, flickering, or ghosting, but continuity is largely preserved. & Primary content is recognizable but degraded by considerable noise, block artifacts, blurring, noticeable flickering, and aliasing. \\
\hline
3 Fair & Minor temporal artifacts; non-severe flickering or ghosting does not majorly impact continuity. & Primary content is clear with some distortions like mild noise, non-severe flickering, visual blurring, or noticeable false edges. \\
\hline
4 Good & Temporal artifacts present but not distracting; continuity well-maintained with minimal flickering. & Clear primary subject with negligible noise or blurring and minor textural or edge distortions, not significantly affecting the experience. \\
\hline
5 Excellent & Excellent temporal stability; no noticeable false edges, flickering, or ghosting. & Primary subject depicted with exceptional clarity, free of distracting distortions, and showing high-quality textural details. \\
\hline \hline
\end{tabular}
\renewcommand{\arraystretch}{1} 
\end{table*}

\subsubsection{Subjective Experiments}\label{sec:SubExp}
In the subjective study, we implement a consistent stimulus evaluation process that allows participants to review the same video multiple times before rating it. The OA-MOS and TS-MOS are annotated on a scale of 1 to 5, with increments of 0.5, following the ITU-R absolute category rating scale \cite{ITU-R, CGF-IQAcomparison}. Quality scores range from 1 ("Bad") to 5 ("Excellent"), with detailed criteria for these ratings provided in Tab.~\ref{tab:mos_criteria}.

The experiments for video quality rating are conducted with 17 trained annotators (10 male and 7 female with normal color vision), all of whom are familiar with rendered content. The research is conducted in a controlled laboratory setting, with each participant required to rate the entire video set on a specific type of display to ensure scoring consistency. The display monitor is color-calibrated to the sRGB standard, with brightness adjusted to 200 cd/$\mathrm{m}^2$ and the white point set to 6500K. The display's refresh rate is adjusted to 60 Hz to match the frame rate of the videos. Desktop monitors are positioned to ensure a comfortable viewing distance of 2 to 3 feet, akin to the experimental setup in \cite{dataset-liveQualcomm}, while the smartphone screen is positioned between 1 and 1.5 feet from the viewer. Three displays are used: a 27" AOC Q27P1U 2K IPS monitor, a 23.8" Dell P2422H 1080P IPS monitor, and a 6.7" AMOLED screen of an OPPO Find X3 Pro smartphone. To facilitate rapid annotation, custom software for both desktop and smartphone platforms has been developed to automate video playback and score recording. These tools can be downloaded from our project homepage.

Before the rating process begins, each participant views an instructional video that details the MOS measures and presents standard examples to illustrate the scoring criteria. During the training session, we specifically emphasize variations in temporal stability to ensure that participants fully understand the concepts of OA-MOS and TS-MOS scoring. Gold standard questions \cite{wu2023exploring} are used to verify annotator performance. Annotators evaluate 10 `golden' videos, and those whose scores significantly diverge from the established standards (a difference greater than 1) are excluded from further stages of the study. All annotators in our tests successfully complete the training sessions, meeting the qualification criteria.

The subjective quality study is divided into three sessions: 720p on a smartphone screen, 1080p on a desktop monitor, and 2K on a desktop monitor. To prevent annotator fatigue, the test process is organized into rounds with rest periods; each round involves evaluating approximately 200 videos over 25 minutes, followed by a 10-minute rest period. Participants may opt to partake in one or more sessions, with compensation provided accordingly. At the conclusion of the subjective experiments, a total of 59,910 video quality ratings are collected from the 17 trained annotators.

\begin{figure}
\centering
\includegraphics[width=\linewidth]{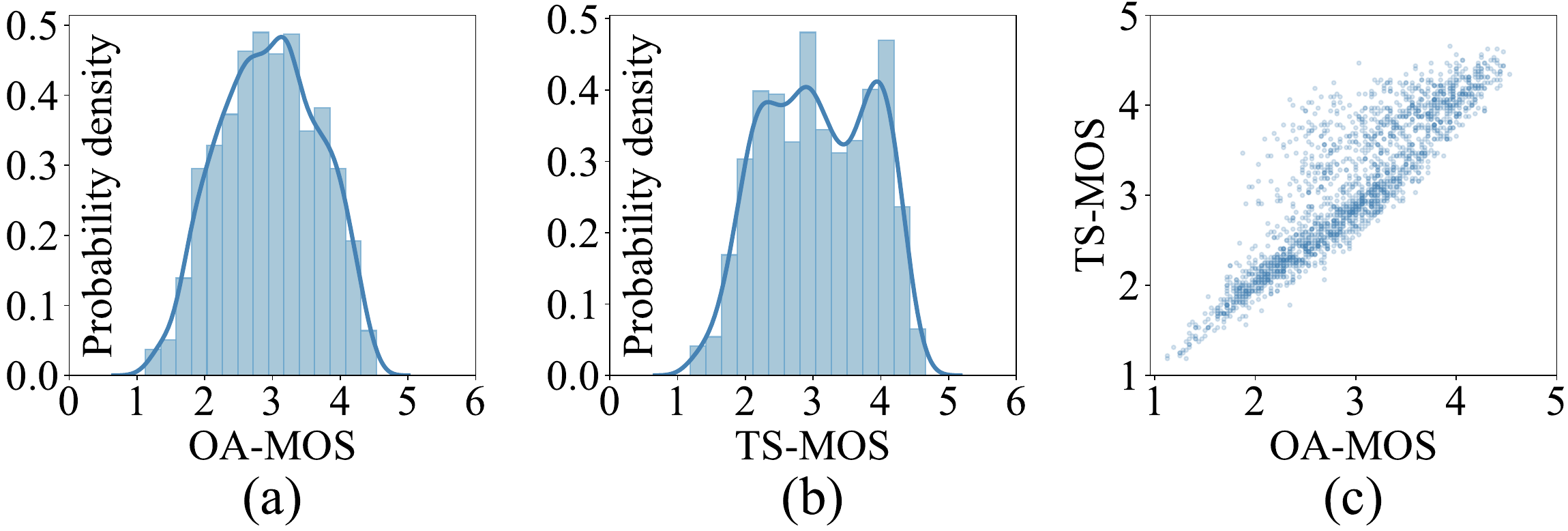}
\caption{Histograms (a) and (b) and scatter plot (c) of OA-MOS and TS-MOS from our proposed ReVQ-2k dataset. The project homepage includes separate data analyses for each resolution and display.
%Separate analyses of data for each resolution and display can be found on the project homepage.
}
\label{fig:MOS}
\end{figure}

% \begin{figure}
% \centering
% \includegraphics[width=\linewidth]{Figs/params.pdf}
% \caption{}
% \label{fig:params}
% \end{figure}

\begin{figure*}
\begin{center}
\includegraphics[width=0.95\linewidth]{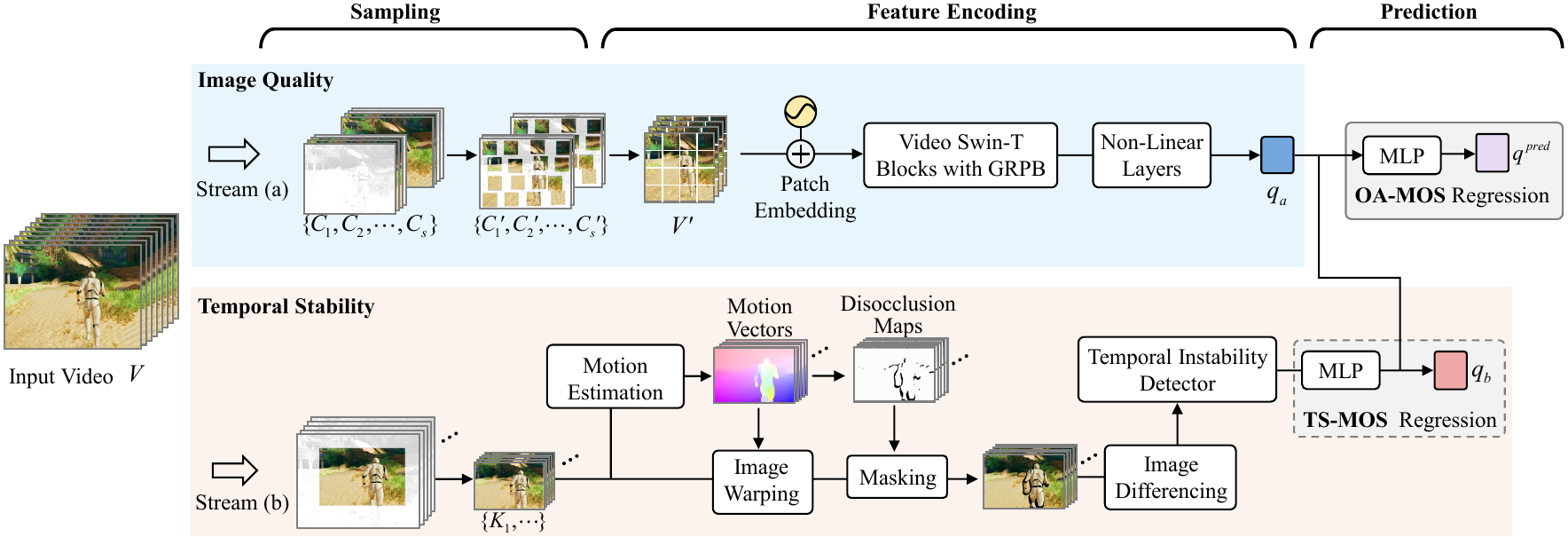}
\end{center}
\caption{Overview of our NR-VQA model. Given an input video \(V\), our model employs two streams, (a) and (b), to extract features relevant to image quality and temporal stability, respectively. Stream (a) utilizes the FAST-VQA framework~\cite{wu2023neighbourhood} to normalize input data and feeds the sampled video \(V'\) into a Swin-T model, which extracts features that are processed through nonlinear layers to derive the image quality score \(q_a\). Stream (b) assesses temporal stability by aligning frames using motion estimation, image warping, and occlusion removal, followed by applying image differencing to compute the temporal score \(q_b\). Both scores, \(q_a\) and \(q_b\), are integrated using an MLP to predict the overall video quality. Notably, stream (b) can be trained using TS-MOS labels to improve the accuracy of NR-VQA for rendered videos.}\label{fig:method_pipeline}
\end{figure*}

\subsubsection{MOS Annotation Analysis} \label{sect:mos_analysis}
\textbf{Data cleaning:} To ensure the validity of the MOS annotations, we implement three data cleaning methods in accordance with ITU-R BT.500-14 \cite{ITU-R}:
1) We reject participants who have at least one annotation score on gold standard videos that deviates by more than 1 unit, as discussed in Sec.~\ref{sec:SubExp}.
2) We include reappearing videos in the annotation process; participants are rejected if the difference in their ratings for any of these videos exceeds 1 unit.
3) We calculate correlation metrics, the Pearson linear correlation coefficient (PLCC) and the Spearman rank-order correlation coefficient (SRCC), between a participant’s annotations and the average scores of other participants. Annotations from any participant whose correlation falls below 0.8 are rejected.
In our experiments, no participants are excluded based on constraints 1) and 2); one participant is excluded due to failing to meet the criteria in constraint 3). From 16 qualified subjects, we have collected 57,450 valid ratings. The averages of the cleaned annotations serve as the MOS results for the videos.

\textbf{Results distribution:} Fig.~\ref{fig:MOS} shows the distribution of TS-MOS and OA-MOS scores. The kurtosis values \cite{kurtosis} for the two distributions in Fig.~\ref{fig:MOS} (a) and (b) are 2.231 and 1.955, respectively. Distributions with kurtosis values below 3 exhibit a plateau shape, indicating a more uniform distribution that captures quality levels across the video dataset effectively, thereby providing a robust basis for algorithm test \cite{dataset-cvd2014}. Fig.~\ref{fig:MOS} (c) shows scatter plots of OA-MOS and TS-MOS, revealing several critical insights. 
First, a large number of videos show a positive correlation between TS-MOS and OA-MOS, implying that videos with good temporal stability have higher overall quality.
% First, a considerable number of the videos show a positive correlation between TS-MOS and OA-MOS, suggesting that videos with good temporal stability generally exhibit higher overall quality.
Second, some data points with low OA-MOS and high TS-MOS suggest that, while these videos maintain good temporal stability, they may suffer from image issues such as blurring or insufficient exposure, lowering their overall quality.
%Second, some data points with low OA-MOS and high TS-MOS indicate that while these videos maintain good temporal stability, they may suffer from image issues such as blurring or inadequate exposure, affecting their overall quality.
Third, there are no data points in the quadrants for high OA-MOS and low TS-MOS. This is primarily because some videos with high image quality rarely exhibit extremely poor temporal stability, whereas some videos with poor temporal stability rarely achieve high perceived quality ratings. These findings emphasize the importance of temporal stability in assessing the quality of rendered videos.
% , and they influenced the development of our new NR-VQA model for rendered content.
% Third, there are no data points in the quadrant representing high OA-MOS and low TS-MOS. This is primarily because videos with high image quality rarely show extremely poor temporal stability, and videos with poor temporal stability rarely achieve high perceived quality ratings. These findings highlight the importance of considering temporal stability in assessing the quality of rendered videos and have influenced the development of our new NR-VQA model tailored for rendered content.
% The probability density distribution of the TS-MOS is more concentrated at both ends, indicating that people are sensitive to flickering in rendered videos. This is consistent with the findings reported in \cite{flicker2008}, which state that severe flickering artifacts negatively affect the perceived quality of videos.
% , with TS-MOS values ranging from 1.08 to 4.75 and OA-MOS values ranging from 1.38 to 4.63, respectively.
% The probability density distribution of the TS-MOS is more concentrated at both ends, indicating that people are sensitive to flickering in rendered videos. This is consistent with the findings reported in \cite{flicker2008}, which state that severe flickering artifacts negatively affect the perceived quality of videos.
\section{Our NR-VQA Method}
\subsection{Overview}
Building on the ReVQ-2k dataset, we develop a new two-stream NR-VQA metric to predict rendered video quality. Fig.~\ref{fig:method_pipeline} illustrates the architecture of our proposed model, which is divided into two main components: image quality assessment and temporal stability analysis. In the image quality assessment stream, we evaluate the overall image quality of videos. Given the extensive analysis in existing literature, we adopt the cropping strategy and Swin transformer (Swin-T) model employed in FAST-VQA \cite{wu2023neighbourhood} (see stream (a) in Fig.~\ref{fig:method_pipeline}). This stream considers factors such as clarity and appropriate exposure, in alignment with existing NR-VQA methods \cite{vsfa19, GSTVQA21}, and evaluates static rendering artifacts like Moiré patterns. In the temporal stability analysis stream (see stream (b) of Fig.~\ref{fig:method_pipeline}), we crop a series of images from consecutive video frames, align them via motion estimation, and assess their temporal stability using image differencing. The results from both streams are then combined through a multilayer perceptron (MLP) to regress the overall video quality, integrating insights from Sec.~\ref{sect:mos_analysis}.

It is worth noting that the entire model can be trained using only the OA-MOS; however, the TS-MOS can be used to train stream (b) prior to the entire model training. An ablation study presented in Sec.~\ref{sec:ab1} shows that using temporal stability evaluations as additional supervision consistently improves prediction accuracy.
% It is noteworthy that the entire model can be trained using only the OA-MOS; alternatively, the TS-MOS can be utilized to train stream (b) specifically before the whole model training. Leveraging temporal stability evaluations as additional supervision has consistently produced steady gains in prediction accuracy, as demonstrated in an ablation study presented in Sec.~\ref{sec:ab1}.

% Most patch locations are randomly selected, although some are specifically aligned temporally to ensure consistency with corresponding grid locations in subsequent frames.  generating fragments \( G_i = \{F'_j, F'_{j+1}, \ldots, F'_{j+m-1}\} \) for each clip \(C_i\). Finally, fragments from all clips are combined

\subsection{Stream (a): Image Quality Evaluation}
For the evaluation of overall image quality (specifically assessing issues such as blurring, noise, overexposure, and static rendering artifacts), extensive research \cite{vsfa19, CNN-TLVQM20, richdatabase2021} has been conducted in the field of NR-VQA for camera-captured videos. Considering the minor gaps in assessing these aspects between rendered and camera-captured videos, we directly employ an existing SOTA NR-VQA practice, FAST-VQA~\cite{wu2023neighbourhood}, for stream (a) of our method. The FAST-VQA method is selected for its well-designed architecture, which has proven to provide efficient and effective assessments of video quality.

As shown in Fig.~\ref{fig:method_pipeline}, our input video \( V\) = \(\{F_1\), \(F_2, \ldots\), \(F_z\} \) consists of \( z \) frames, with \( F_i \) denoting the \( i \)-th frame. In stream (a), the video is segmented into \( s \) clips, each containing \( t = z/s \) consecutive frames. For each clip \( C_i = \{F_{i \cdot t}, F_{i \cdot t + 1}, \ldots, F_{i \cdot t + t - 1}\} \), we randomly select \( m \) consecutive frames to form a subset \( C'_i = \{F_j, F_{j+1}, \ldots, F_{j+m-1}\} \), where \( j \) is randomly chosen from (\( i \cdot t \)) to (\( i \cdot t + t - m \)). Each frame \( F_j \) in \( C'_i \) is then divided into an \( n \times n \) grid; then, a \( k \times k \) image patch is cropped from each grid cell. The cropped patches are concatenated to form an image \( F'_i \) of dimensions \((n \cdot k) \times (n \cdot k)\). This procedure is replicated for all images in image subsets \( \{C'_1, C'_2, \ldots, C'_s\} \), producing a resampled video \( V'\). This approach utilizes key parameters: \( s = 8 \), \( m = 4 \), \( n = 7 \), and \( k = 32 \). Such a sampling strategy not only significantly reduces data volume by eliminating potential redundancy but also standardizes input videos of varying lengths and resolutions into a uniform format without the need for resizing. Readers are referred to FAST-VQA \cite{wu2023neighbourhood} for further details.

After the sampling process, the video \( V' \) undergoes patch embedding and is processed using a Swin-T model to extract high-level features related to video quality assessment. It is important to note that \( V' \) is constructed from small patches, which can introduce discontinuities at the seams of patches. Consequently, the feature extraction network must be carefully designed to avoid misinterpreting these seams as image artifacts. To address this, FAST-VQA \cite{wu2023neighbourhood} employs non-overlapping pooling kernels, effectively preventing the misinterpretation of patch seams as artifacts. Additionally, FAST-VQA introduces gated relative position biases (GRPB) within the Swin-T to enhance the performance of the self-attention layers. Finally, the features extracted from each patch are processed through non-linear layers to derive a quality score \( q_a \), which assesses the image quality of the input video.

\begin{figure}
\begin{center}
\includegraphics[width=\linewidth]{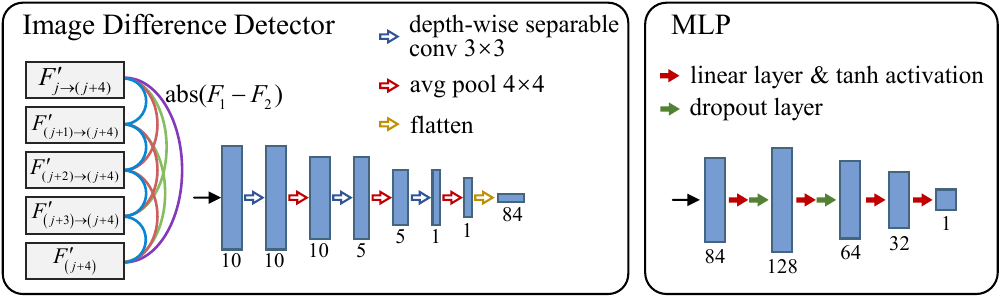}
\end{center}
\caption{Detailed structures of the image difference detector and the MLP model. The numbers in the diagram represent the feature channels.}
\label{fig:streambc}
\end{figure}

% In contrast to the image quality assessment task, information on temporal dimension is important in the video quality assessment problem, as demonstrated in previous work \cite{vsfa19, CNN-TLVQM20, richdatabase2021}. However, these frame-based methods are incapable of dealing with abrupt changes in the temporal domain. While the optical flow contains a wealth of statistics that can be modeled at the frame-pair level. We use the optical flow estimation in our neural network to distill video features in the temporal domain, inspired by the application of optical flow in video recognition \cite{opticalflowvideorecognition}. It allows the model to extract more accurate temporal information.

\subsection{Stream (b): Temporal Stability Evaluation}
While some existing NR-VQA methods have started to incorporate temporal stability into video quality assessment, they often lack the capability to effectively detect temporal artifacts such as flickering and moving jaggies, which are rarely present in camera-captured content.
% Existing NR-VQA methods generally do not account for temporal artifacts such as flickering and moving jaggies in assessing video quality, as these issues are uncommon in camera-captured content. 
However, in rendered videos, temporal artifacts are prevalent and significantly impact video quality~\cite{yang2020survey}. Although FAST-VQA employs temporally contiguous image patches in video sampling, it fails to detect instability due to the common misalignment of moving objects in videos. To address this, we propose a temporal artifacts detector that forms stream (b) of our metric. This stream uses motion estimation to align objects across frames and utilizes image differencing to detect pixel-level temporal instability, thereby playing a vital role in the assessment of rendered video quality.

As depicted in stream (b) of Fig.~\ref{fig:method_pipeline}, to reduce computational overhead and standardize input data shape, we first sample the input video into a normalized shape. We uniformly extract 10 subsets from the input video \( V \), each containing 5 consecutive frames. To normalize the resolution, these subsets are cropped to a uniform resolution of \(h = 480, w = 800\), producing standardized subsets \( \{K_1, K_2, \ldots, K_{10}\} \). The cropping location is consistent across all frames within each subset but is randomly selected relative to the original frame positions. Each subset \( K_i \) then undergoes motion estimation, generating motion vectors \( \{M_{j}, M_{j+1}, M_{j+2}, M_{j+3}\} \) for frames \( \{F'_{j}, F'_{j+1}, F'_{j+2}, F'_{j+3}\} \) to \( F'_{j+4} \), where \( F'_{j} \) represents the cropped image of frame \( F_{j} \). We employ the SOTA motion estimation algorithm, dense optical tracking (DOT) \cite{lemoing2024dense}, to efficiently track points across video frames by capturing key motion tracks from dynamic boundaries.
% which efficiently tracks points across video frames by capturing key motion tracks from dynamic boundaries. 
Additionally, we have enhanced the DOT algorithm to mark occluded areas, enabling the concurrent generation of disocclusion maps \( \{D_{j}, D_{j+1}, D_{j+2}, D_{j+3}\} \). These maps identify regions in \( F'_{j+4} \) where objects in \( \{F'_{j}, F'_{j+1}, F'_{j+2}, F'_{j+3}\} \) are occluded, effectively marking pixels lacking temporal correspondences. After generating motion vectors and disocclusion maps, each subset \( K_i \) is subjected to backward warping and disoccluded pixel removal, resulting in an aligned subset \( K'_i = \{F'_{j\rightarrow(j+4)}, F'_{(j+1)\rightarrow(j+4)}, \ldots, F'_{j+4}\} \), with \( F'_{j\rightarrow(j+4)} \) obtained by:
\begin{equation}
\label{eq:rgb_frame_feature}
F'_{j\rightarrow(j+4)} = \text{Warping}(F_j, M_j) \cdot D,
\end{equation}
where \(D\) represents the overlap of the four disocclusion maps.
Then, the processed image subsets \(\{K'_1, K'_2, \ldots, K'_{10}\} \) undergo image differencing over various time spans. As shown in Fig.~\ref{fig:streambc}, we calculate image differencing for adjacent frames and for frames separated by one, two, and three intervals, enabling detection of flickering across various frequencies. The image differences are then fed into a depth-wise separable convolutions-based \cite{howard2017mobilenets} image difference detector to evaluate pixel-level temporal stability. Finally, the stability maps from all subsets are subjected to average pooling and an MLP to regress the temporal stability score \( q_b \). The detailed structures of the image difference detector and the MLP are presented in Fig.~\ref{fig:streambc}.

% , which is analyzed by a MLP to assess the temporal stability \( Q_{\text{TS}} \) of the input video, as shown in Fig.~\ref{fig:method_pipeline}.

% It worth noting that Stream (c) can be trained end-to-end globally, as well as being trained specifically with the TS-MOS labels and subsequently frozen.

% This MLP comprises five fully connected (FC) layers with output channels [256, 128, 64, 32, 1] and utilizes the Leaky ReLU activation function.

\subsection{Final MOS Prediction}\label{sec:final}
After obtaining the image quality score \( q_a \) (which can be seen as the result of the FAST-VQA \cite{wu2023neighbourhood} method) and temporal stability score \( q_b \), we aim to determine their relationship to the overall video quality, expressed as \( q^{pred} = f(q_a, q_b) \). As discussed in Sec.~\ref{sect:mos_analysis}, there exists a non-linear relationship between the temporal stability of rendered videos and their overall perceived quality. Here, we implement an MLP to model the complex non-linear mapping between (\( q_a \), \( q_b \)) and \( q^{pred} \). This MLP utilizes a configuration similar to that used in stream (b), but with only a quarter of the number of neurons to prevent overfitting. Experimental results presented in Sec.~\ref{sec:ab1} also demonstrate that this non-linear mapping approach achieves superior accuracy compared to linear mapping methods.

For the loss functions, extensive exploration has been conducted in prior studies~\cite{sun2022deep, wu2023exploring}, with research primarily focusing on the correlation between predicted scores and ground truth. Consistent with prevalent practices, we adopt the widely used PLCC and ranking loss functions. Given a batch of predicted quality scores \( Q^{pred} = \{q^{pred}_1, q^{pred}_2, \ldots, q^{pred}_s\} \) and ground truth labels \( Q = \{q_1, q_2, \ldots, q_s\} \), these loss functions are defined as:
\begin{equation}
\begin{aligned}
&L_{\text{PLCC}} = \left ( 1-\frac{\sum_{i=1}^s \left ( q^{pred}_i - a\right )\left ( q_i - b \right )}{\sqrt{\sum_{i=1}^s ( q^{pred}_i - a )^2 \sum_{i=1}^s ( q_i - b )^2}} \right ) / 2,\\
&L_{\text{ranking}} = \frac{1}{s^2} \sum_{i=1}^s \sum_{j=1}^s \text{max} \left ((q^{pred}_j - q^{pred}_i) \text{sgn}(q_i-q_j), 0 \right ),
\end{aligned}
\end{equation}
where \( a \) and \( b \) are the mean values of \( Q^{pred} \) and \( Q \), respectively, and \( \text{sgn}(\cdot) \) denotes the sign function. We empirically set a weight \(\alpha\) to 0.3 and combine the loss functions as:
\begin{equation}
\text{LOSS} = L_{\text{PLCC}} + \alpha \cdot L_{\text{ranking}},
\end{equation}
which is used to train both stream (b) and our entire model.

% \end{equation}

\begin{table*}[]
\caption{Quantitative comparison of NR-VQA methods on the ReVQ-2k dataset (720p, 1080p, and 2k resolutions). Metrics shown include PLCC and SRCC. \textbf{Boldface} denotes the best performance and \underline{underline} indicates the second-best performance for each metric.}\label{tab:res1}
\centering\renewcommand{\arraystretch}{1.2} 
\footnotesize
\begin{tabular}{>{\raggedleft\arraybackslash}p{2.4cm}|>{\centering\arraybackslash}p{1.2cm} >{\centering\arraybackslash}p{1.2cm}|>{\centering\arraybackslash}p{1.2cm} >{\centering\arraybackslash}p{1.2cm}|>{\centering\arraybackslash}p{1.2cm} >{\centering\arraybackslash}p{1.2cm}|>{\centering\arraybackslash}p{1.2cm} >{\centering\arraybackslash}p{1.2cm}}
\hline
\hline
\multirow{2}{*}{Methods} & \multicolumn{2}{c|}{ReVQ-2k (720p)} & \multicolumn{2}{c|}{ReVQ-2k (1080p)} & \multicolumn{2}{c|}{ReVQ-2k (2k)} & \multicolumn{2}{c}{Weighted Average} \\
                                    & SRCC$\uparrow$  & PLCC$\uparrow$  & SRCC$\uparrow$  & PLCC$\uparrow$  & SRCC$\uparrow$  & PLCC$\uparrow$  & SRCC$\uparrow$  & PLCC$\uparrow$ \\ \hline
VSFA~\cite{vsfa19}                  & 0.785  & 0.767  & 0.755  & 0.738  & 0.772  & 0.780  & 0.771  & 0.762 \\
CNN-TLVQM~\cite{CNN-TLVQM20}        & 0.782  & 0.784  & 0.773  & 0.774  & 0.762  & 0.779  & 0.772  & 0.779 \\
GSTVQA~\cite{GSTVQA21}              & 0.839  & 0.829  & 0.836  & 0.817  & 0.826  & 0.822  & 0.834  & 0.823 \\
SimpleVQA~\cite{sun2022deep}        & 0.801  & 0.808  & 0.759  & 0.785  & 0.780  & 0.796  & 0.780  & 0.796 \\
DOVER~\cite{wu2023exploring}        & 0.821  & 0.829  & 0.809  & 0.817  & 0.812  & 0.815  & 0.814  & 0.820 \\
FAST-VQA~\cite{wu2023neighbourhood} & 0.846  & 0.845  & 0.852  & 0.856  & 0.823  & 0.849  & 0.840  & 0.850 \\
MBVQA~\cite{wen2024modular}         & 0.818  & 0.826  & 0.803  & 0.821  & 0.814  & 0.839  & 0.812  & 0.829 \\ \hline
Ours-                               & \underline{0.854}  & \underline{0.859}  & \underline{0.864}  & \underline{0.863}  & \underline{0.836}  & \underline{0.853}  & \underline{0.851}  & \underline{0.858} \\
Ours                                & \textbf{0.882}  & \textbf{0.884}  & \textbf{0.885}  & \textbf{0.887}  & \textbf{0.869}  & \textbf{0.874}  & \textbf{0.879}  & \textbf{0.882} \\
\hline \hline
\end{tabular}
\end{table*}

\section{Experiments}
\label{Exp}
This section describes how we implemented our method and compares it to existing VQA methods on both the proposed ReVQ-2k dataset and the existing NR-VQA dataset. Then we perform ablation studies to determine the impact of individual components in our model.
% This section details the implementation of our method and presents comparative evaluations with existing VQA methods on both the proposed ReVQ-2k dataset and existing NR-VQA datasets. Finally, we conduct ablation studies to assess the impact of individual components within our model.

%In this section, implementation details of the experiments will be first introduced, including competing methods, test datasets, and performance criteria. Then we will compare our model with other under current state-of-the-art framework both on in-the-wild video datasets and rendering-content video dataset. We also conduct cross dataset experiments to verify the generalization capability of our method. Finally, the ablation study is introduced.

% To reduce training time, the data are pre-sampled and motion vectors are computed via motion estimation. 

\subsection{Implementation Details}\label{Implementation}
In our method, stream (b) can be independently trained using TS-MOS labels or integrated with the entire model trained using only OA-MOS labels. We implement both strategies using a batch size of 16 and the Adam optimizer with a learning rate of 0.001. The remaining training configurations are maintained consistent with those used in the FAST-VQA method. Given that the motion estimation phase is typically more time-consuming compared to neural network inference, we recommend precomputing motion vectors for the dataset to reduce both training and test times. Our experiments are conducted on a desktop PC equipped with an NVIDIA GeForce RTX 4090 GPU, an Intel i7-13700K CPU, and 64GB of memory.

\subsection{Evaluation Setups}
\textbf{Datasets.}
To evaluate the performance of NR-VQA methods, we utilize the proposed ReVQ-2k dataset, along with established NR-VQA datasets\footnote{We did not conduct experiments on the largest NR-VQA dataset, LSVQ~\cite{patchvq}, for two main reasons. First, the dataset’s extensive video collection would demand an impractically large amount of time for motion estimation of our method. Second, the dataset primarily consists of camera-captured videos, which are not directly relevant for evaluating the performance of our rendering-oriented method.}, including CVD2014 \cite{dataset-cvd2014}, LIVE-Qualcomm \cite{dataset-liveQualcomm}, KoNViD-1k \cite{kon-1kdataset}, LIVE-VQC \cite{dataset-LIVEVQC}, and LIVE-YT-Gaming \cite{yu2023subjective}. Detailed descriptions of these datasets are presented in Tab.~\ref{table:datasets}. The ReVQ-2k dataset includes videos at three resolutions: 2K, 1080P, and 720P, each played and annotated on their respective displays. Therefore, comparisons on the ReVQ-2k dataset are performed separately for each resolution. Note that we consider videos from each 3D scene as unique entities for the training and test phases, ensuring scene independence during the training/test set splitting.
% ; videos from different scenes will not appear in the same training batch. 
During the test, results from each scene are independently calculated and then combined to produce a weighted average. This approach addresses potential biases due to varying scene content, more closely reflecting real-world applications. 
% Additionally, we extend our evaluations to existing NR-VQA datasets, which primarily consist of camera-captured videos.

% To ensure fairness, our approach is trained end-to-end on these datasets without using any modules pretrained on ReVQ-2k, consistent with existing methods.
% This extension enables us to ascertain whether our spatial and temporal artifact detection methods are also effective in improving the video quality assessment of camera-captured content.

% \footnote{We did not conduct experiments on the largest NR-VQA dataset, LSVQ~\cite{patchvq}, due to two main reasons. First, the extensive number of videos in LSVQ would require several weeks for motion estimation. Second, the dataset primarily includes camera-captured videos at various resolutions, which are less relevant to our study focusing on rendered video quality assessment. Therefore, we decided not to invest resources into validating our method on this dataset.}

% \noindent

\textbf{Evaluation Metrics.}
We assess NR-VQA models using two commonly used criteria: the PLCC and the SRCC. The PLCC is employed to evaluate the accuracy of predictions by measuring the linear relationship between predicted and actual values. The SRCC is used to assess the monotonicity of predictions, evaluating how consistently the predictions preserve the ordinality of the actual values. When addressing datasets with varying MOS scales, we utilize a logistic function \(g\), as suggested by Li \textit{et al}.~\cite{vsfa19}, to map the predicted scores \(o\) to the corresponding subjective scores \(s\):
\begin{equation}
\label{eq:score_mapping}
     g(o) = \frac{\beta_{1} - \beta_{2}}{1 + e^{-\frac{o - \beta_{3}}{\beta_4}}} + \beta_{2},
\end{equation}
where \(\beta_1 = \max(s)\),
\(\beta_2 = \min(s)\),
\(\beta_3 = \text{mean}(o)\),
and \(\beta_4 = \text{std}(o)/4\).

% \begin{figure*}
% \centering
% \includegraphics[width=\linewidth]{Figs/2scenes.png}
% \includegraphics[width=\linewidth]{Figs/2scenes.png}
% \caption{Visual comparison of video quality predictions from various NR-VQA methods on the ReVQ-2k dataset. Temporal profiles of thin bars illustrate the temporal stability assessment, highlighting our method's ability to capture fine temporal variations compared to baselines.}
% \label{fig:res1}
% \end{figure*}

\begin{figure*}
\centering
\includegraphics[width=0.95\linewidth]{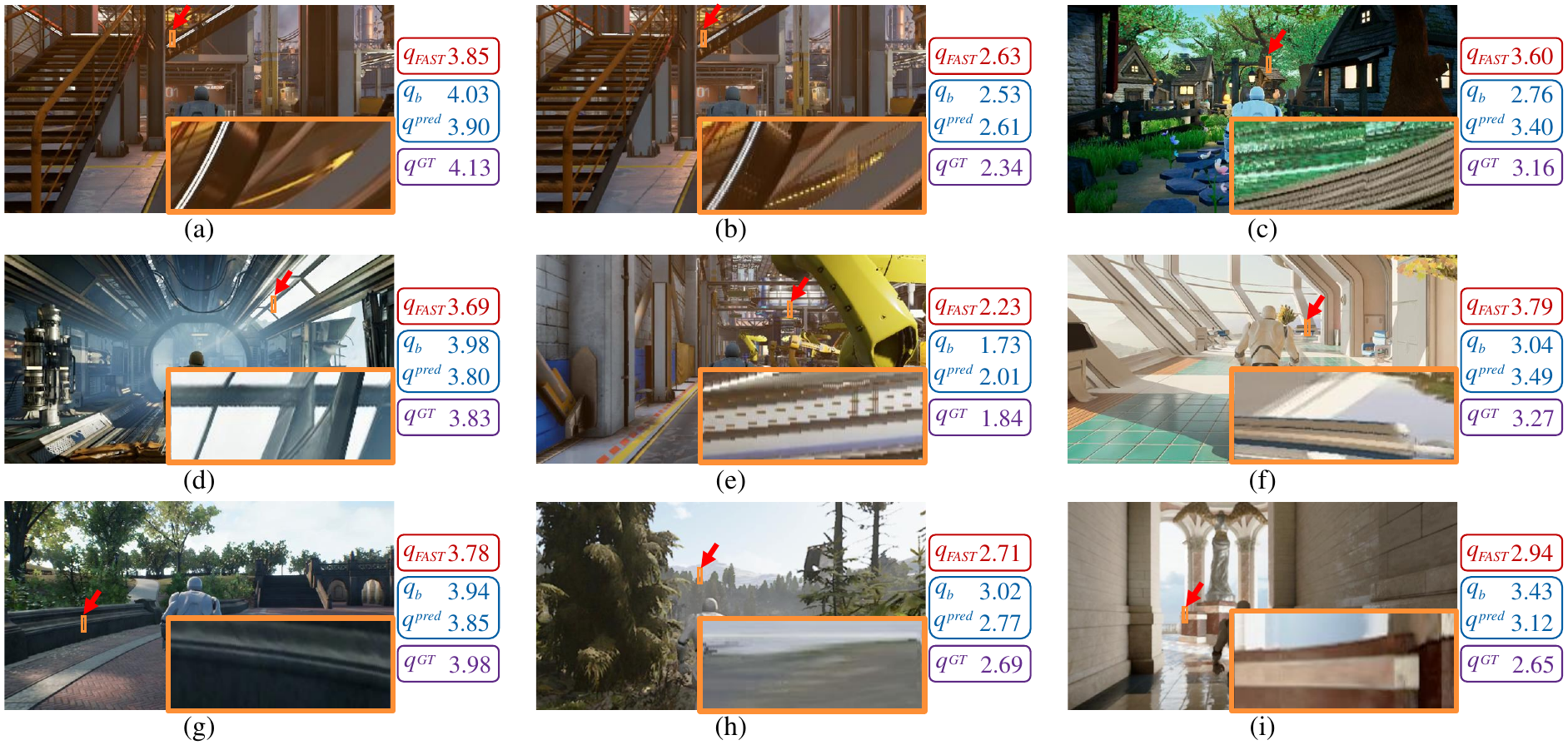}
\caption{Visual comparison of video quality predictions by FAST-VQA and our method on the ReVQ-2k dataset. Temporal profiles (orange boxes) are depicted through a column of pixels across temporal frames. \(q_{FAST}\) represents predictions from FAST-VQA, while \(q_b\), \(q^{pred}\), and \(q^{GT}\) denote the results from stream (b), our final result, and the ground truth video quality (OA-MOS), respectively. The predictions are rescaled using the mean and standard deviation of the ground truth annotations.
% All results are rescaled to match the range of the ground truth annotations. 
This figure illustrates how our method integrates image quality (using \(q_{FAST}\) as a reference) with temporal stability assessments (\(q_b\)) to achieve a comprehensive evaluation of rendered video quality (\(q^{pred}\)) .}
\label{fig:res1}
\end{figure*}

\begin{table*}[]\label{tab:res2}
\caption{Comparison of NR-VQA methods on established datasets including CVD2014, LIVE-Qualcomm, KoNViD-1k, LIVE-VQC, and LIVE-YT-Gaming.}
\centering\renewcommand{\arraystretch}{1.2} 
\footnotesize
\begin{tabular}{>{\raggedleft\arraybackslash}p{2.4cm}|>{\centering\arraybackslash}p{1.1cm} >{\centering\arraybackslash}p{1.1cm}|>{\centering\arraybackslash}p{1.1cm} >{\centering\arraybackslash}p{1.1cm}|>{\centering\arraybackslash}p{1.1cm} >{\centering\arraybackslash}p{1.1cm}|>{\centering\arraybackslash}p{1.1cm} >{\centering\arraybackslash}p{1.1cm}|>{\centering\arraybackslash}p{1.1cm} >{\centering\arraybackslash}p{1.1cm}}
\hline
\hline
\multirow{2}{*}{Methods} & \multicolumn{2}{c|}{CVD2014} & \multicolumn{2}{c|}{LIVE-Qualcomm} & \multicolumn{2}{c|}{KoNViD-1k} & \multicolumn{2}{c|}{LIVE-VQC}  & \multicolumn{2}{c}{LIVE-YT-Gaming} \\
                                    & SRCC$\uparrow$  & PLCC$\uparrow$  & SRCC$\uparrow$  & PLCC$\uparrow$  & SRCC$\uparrow$  & PLCC$\uparrow$  & SRCC$\uparrow$  & PLCC$\uparrow$ & SRCC$\uparrow$  & PLCC$\uparrow$ \\ \hline
VSFA~\cite{vsfa19}                  & 0.850  & 0.869  & 0.708  & 0.774  & 0.794  & 0.799  & 0.718  & 0.771 & 0.784  & 0.819 \\
CNN-TLVQM~\cite{CNN-TLVQM20}        & 0.863  & 0.880  & 0.810  & \underline{0.833}  & 0.816  & 0.818  & 0.825  & 0.834 & 0.855  & 0.866 \\
GSTVQA~\cite{GSTVQA21}              & 0.831  & 0.844  & 0.801  & 0.825  & 0.814  & 0.825  & 0.788  & 0.796 & 0.850  & 0.860 \\
SimpleVQA~\cite{sun2022deep}        & 0.834  & 0.864  & 0.722  & 0.774  & 0.792  & 0.798  & 0.740  & 0.775 & 0.814  & 0.836 \\
DOVER~\cite{wu2023exploring}        & 0.858  & 0.881  & 0.736  & 0.789  & 0.892  & \underline{0.900}  & 0.853  & 0.872 & \underline{0.882}  & \textbf{0.906} \\
FAST-VQA~\cite{wu2023neighbourhood} & \textbf{0.883}  & \textbf{0.901}  & 0.807  & 0.814  & 0.893  & 0.887  & 0.853  & \underline{0.873} & 0.869  & 0.880 \\
MBVQA~\cite{wen2024modular}         & \textbf{0.883}  & \textbf{0.901}  & \textbf{0.832}  & \textbf{0.842}  & \textbf{0.901}  & \textbf{0.905}  & \textbf{0.860}  & \textbf{0.880} & 0.867  & 0.902 \\ \hline
Ours-                               & \underline{0.881}  & \underline{0.895}  & \underline{0.816}  & 0.828  & \underline{0.896}  & 0.894  & \underline{0.857}  & 0.869 & \textbf{0.891}  & \underline{0.904} \\
\hline
\hline
\end{tabular}
\end{table*}

\subsection{Comparisons on Rendered Video Dataset}
We compare our method against existing SOTA baselines, including VSFA~\cite{vsfa19}, CNN-TLVQM~\cite{CNN-TLVQM20}, GSTVQA~\cite{GSTVQA21}, SimpleVQA~\cite{sun2022deep}, DOVER~\cite{wu2023exploring}, FAST-VQA~\cite{wu2023neighbourhood}, and the full version of MBVQA~\cite{wen2024modular}. To assess their performance on the ReVQ-2k dataset, we utilize the source code provided by the authors, training their models under the recommended settings to ensure optimal performance. All methods, including ours, are initially pretrained on the training set of the large-scale LSVQ dataset \cite{patchvq}, and then fine-tuned on the test datasets. We implement our approach in two variants: one trained solely with OA-MOS labels, denoted as ``Ours-'', and another that incorporates TS-MOS labels for additional supervision, denoted as ``Ours''. To minimize variability due to random dataset partitioning, we repeatedly split the 3D scenes into training and test sets five times, maintaining a splitting ratio of approximately 8:2 for training and test. The specifics of these splits are included in the public release of the dataset.

% Tab.~\ref{tab:res1} and Fig.~\ref{fig:res1} present the results of our method and compare it against the best-performing baselines on the ReVQ-2k dataset. 
The quantitative analysis in Tab.~\ref{tab:res1} clearly demonstrates that our proposed model significantly outperforms existing SOTA methods. Without temporal stability supervision, our model (Ours-) exceeds the top-performing baseline by 1.1\% and 0.8\% in PLCC and SRCC, respectively. This margin increases to 3.9\% and 3.2\% when our model (Ours) incorporates temporal stability scores for training. While the DOVER method~\cite{wu2023exploring}, which incorporates aesthetic opinions, proves effective for camera-captured videos, using a pretrained aesthetic evaluation module offers no improvement in NR-VQA performance on rendered datasets. Similarly, the MBVQA method~\cite{wen2024modular}, introducing spatial and temporal rectifiers, also fails to deliver advantages for rendered data.

Fig.~\ref{fig:res1} illustrates the visual results with temporal profiles of a column of pixels across temporal frames, effectively visualizing the temporal stability of videos. Notably, since FAST-VQA does not directly evaluate video temporal stability, there are frequent discrepancies between its predictions (\(q_{FAST}\)) and the ground truth (\(q^{GT}\)). For instance, in panels (a), (d), and (g), where videos exhibit good temporal stability, FAST-VQA often underestimates the overall quality; conversely, in panels (b), (c), (e), and (f), which are characterized by poor temporal stability, it yields overly high quality estimates. By integrating the temporal stability evaluation \(q_b\), we enhance the assessment of temporal artifacts, thus aligning the results \(q^{{pred}}\) more closely with human-annotated scores. This analysis demonstrates the effectiveness of incorporating temporal stability assessments in NR-VQA for rendered videos.

\subsection{Comparisons on Existing NR-VQA Datasets}
We also test our method on existing datasets of camera-captured content, despite these tests are not directly related to the evaluation of our metric designed for rendered videos. For the CVD2014~\cite{dataset-cvd2014}, LIVE-Qualcomm~\cite{dataset-liveQualcomm}, KoNViD-1k~\cite{kon-1kdataset}, LIVE-VQC \cite{dataset-LIVEVQC}, and LIVE-YT-Gaming~\cite{yu2023subjective} datasets, we adhere to established research protocols by conducting experiments using 10 random train-test splits, allocating 80\% of the data for training and 20\% for test. All methods are pretrained on the training set of the large-scale LSVQ dataset \cite{patchvq}. Because these datasets lack TS-MOS annotations, we only present the ``Ours-" model, which is trained with only OA-MOS labels.
% As these datasets do not include TS-MOS annotations, we only present the ``Ours-" model that is trained using only OA-MOS labels. 
The comparative results, shown in Tab.~\ref{tab:res2}, demonstrate that our method is competitive in assessing camera-captured video quality. For most datasets, our method achieves top-2 performance in both PLCC and SRCC scores. Although the improvements compared to baselines on these datasets are not as pronounced as those observed on the rendered video dataset, our model still exhibits strong robustness. Specifically, our method shows superior performance on the LIVE-YT-Gaming dataset, showing potential future extensions to NR-VQA for cloud gaming and streaming games.

\subsection{Ablation Study}
\subsubsection{Effectiveness of Temporal Stability Evaluation} \label{sec:ab1}
We evaluate the contributions of streams (b), TS-MOS annotations, and non-linear MOS regression within our model through seven variants: 1) using only stream (a); 2) using only stream (b), trained and evaluated on OA-MOS labels; 3) using only stream (b), trained on TS-MOS labels but evaluated on OA-MOS labels; 4) combining streams (a) and (b) with a linear function; 5) combining streams (a) and (b) with a non-linear function; 6) incorporating TS-MOS training into variant 4; and 7) incorporating TS-MOS training into variant 5. The tests are carried out on the ReVQ-2k dataset (720p and 1080p) using five train-test splits. The results, detailed in Tab.~\ref{tab:ab1}, demonstrate significant enhancements in video quality prediction accuracy with the integration of temporal stability evaluation by streams (b) and the TS-MOS annotations. Including stream (b) and using TS-MOS labels for training results in average improvements of 4.5\% in SRCC and 4.0\% in PLCC, and using non-linear for MOS regression further enhances the results. 
% This experiment demonstrates the importance of incorporating temporal stability in NR-VQA for rendered videos and shows that using the non-linear function for improved performance.

\subsubsection{Effectiveness of Motion Estimation and Masking}
To effectively assess temporal stability in videos, our method integrates motion estimation and occlusion masking in stream (b). We conduct experiments to validate the effectiveness of these components by comparing three model configurations: one without motion estimation and masking, one with motion estimation but no masking, and our final model with both.\footnote{Note that only stream (b) is tested, with TS-MOS labels used for both training and test. }
The results, reported in Tab.~\ref{tab:ab2}, reveal that the absence of motion estimation significantly compromises the model’s accuracy in predicting temporal stability, with an SRCC of 0.170 and a PLCC of 0.163. Introducing motion estimation alone improves these metrics to 0.782 and 0.788, respectively. Our complete model, incorporating both motion estimation and masking, achieves further improvements, attaining an SRCC of 0.842 and a PLCC of 0.853. These findings validate the effectiveness of incorporating motion estimation and masking for enhanced temporal stability assessment.

\begin{table}
\caption{Comparison of video quality prediction accuracy with streams (b), TS-MOS labels, and the MLP model on the ReVQ-2k dataset.}
\centering
\centering\renewcommand{\arraystretch}{1.2} 
\footnotesize
\begin{tabular}{l|cc|cc} 
\hline\hline
\multirow{2}{*}{Methods} & \multicolumn{2}{c|}{ReVQ-2k (720p)} & \multicolumn{2}{c}{ReVQ-2k (1080p)} \\
                & SRCC$\uparrow$  & PLCC$\uparrow$  & SRCC$\uparrow$  & PLCC$\uparrow$  \\ \hline 
\( q_a \)                                & 0.843  & 0.840  & 0.825  & 0.836 \\
\( q_b \)                                & 0.728  & 0.713  & 0.753  & 0.768 \\
\( q_b \) \& TS-MOS                      & 0.683  & 0.695  & 0.721  & 0.729 \\\hline 
Lin\(\left( q_a, q_b \right)\)           & 0.848  & 0.858  & 0.859  & 0.861 \\
MLP\(\left( q_a, q_b \right)\)           & 0.854  & 0.859  & 0.864  & 0.863 \\
Lin\(\left( q_a, q_b \right)\) \& TS-MOS & 0.876  & 0.880  & 0.881  & 0.876 \\
MLP\(\left( q_a, q_b \right)\) \& TS-MOS & \textbf{0.882}  & \textbf{0.884}  & \textbf{0.885}  & \textbf{0.887} \\
\hline\hline
\end{tabular}

\label{tab:ab1}
\end{table}

\begin{table}
\caption{Comparison of motion estimation and masking on temporal stability prediction, with results evaluated by TS-MOS labels.}
\centering
\centering\renewcommand{\arraystretch}{1.2} 
\footnotesize
\begin{tabular}{l|cc|cc} 
\hline\hline
\multirow{2}{*}{Methods} & \multicolumn{2}{c|}{ReVQ-2k (720p)} & \multicolumn{2}{c}{ReVQ-2k (1080p)} \\
                & SRCC$\uparrow$  & PLCC$\uparrow$  & SRCC$\uparrow$  & PLCC$\uparrow$  \\ \hline
\textit{\textbf{w/o}} motion \& masking                     & 0.187  & 0.214  & 0.152  & 0.113 \\
\textit{\textbf{w}} motion, \textit{\textbf{w/o}} masking   & 0.779  & 0.786  & 0.785  & 0.790 \\
\textit{\textbf{w}} motion \& masking                       & \textbf{0.831}  & \textbf{0.856}  & \textbf{0.852}  & \textbf{0.850} \\
\hline\hline
\end{tabular}

\label{tab:ab2}
\end{table}

\section{Applications}
This section presents two real-world applications of our NR-VQA metric. First, we apply the metric to assess video quality across various closed-source supersampling methods for mobile real-time rendering, addressing scenarios where videos cannot be perfectly aligned. Second, we utilize the metric to evaluate the perceived quality of various frame generation strategies for real-time rendering. These applications demonstrate the practical utility of our NR-VQA metric in rendering program development.

\subsection{Benchmarking Mobile Supersampling Methods}
In the real-time rendering, supersampling is extensively employed to achieve anti-aliasing or super-resolution (SR) in images, with its application rapidly expanding in mobile real-time rendering. Lightweight supersampling methods suitable for mobile platforms include hardware-dependent solutions such as Snapdragon's game SR~\cite{qualcommSR} and Pixelworks' hardware SR method~\cite{pixelworks}, as well as hardware-independent methods like AMD FSR 1.0~\cite{fsr1} and 2.0~\cite{fsr2}, and MNSS~\cite{yang2023mnss}. Additionally, our team has recently developed a new lightweight SR method~\cite{yang2024frame}. However, comparing our SR method to existing ones is challenging due to misalignment issues in videos produced by various closed-source SR algorithms, where object positions and character poses cannot be consistently aligned across videos. This misalignment makes full-reference methods such as SSIM~\cite{wang2004image} and VMAF~\cite{VMAF24} impractical. In constrast, our NR-VQA method enables effective quality evaluation of these non-aligned videos.
% , making it ideal for assessing the performance of various SR methods.

To evaluate the performance of different SR methods, we collect videos from two 3D game scenes, each with $\times 2$ resolution upscaling. Each method generates five 8-second videos at 60 FPS / 720p for each scene. To minimize quality biases caused by content variations, efforts are made to ensure that the video content from different SR methods is as similar as possible. We employ our NR-VQA metric, calibrated on the ReVQ-2k (720p) dataset, to perform perceptual video quality scoring. For commercial considerations, we anonymize the names of the SR methods, using codes $O_1, O_2, O_3$ for versions of our SR model, and $A_1, A_2, A_3, A_4$ for comparative methods. To ensure a fair and consistent evaluation, all videos are annotated by the trained participants from the proposed ReVQ-2k dataset to derive the reference OA-MOS scores.

The experiment results, as reported in Tab.~\ref{tab:app1}, indicate that method $A_4$ achieves the highest quality scores across all scenes. The results for $O_3$ are comparable to those of $A_3$, while $A_1$ and $A_2$ underperform. 
% We report the detailed experimental results in Tab.~\ref{tab:app1}, including the predicted quality scores of our method and the reference OA-MOS labels.
% Additionally, we conduct subjective studies to obtain MOS scores, as reported in the `OA-MOS' columns of Tab.~\ref{tab:app1}. 
Additionally, the strong correlation between our model’s predicted video quality and the manually annotated OA-MOS confirms the accuracy of our automated method. The proposed NR-VQA metric facilitates immediate quantitative evaluations of our SR algorithm and other closed-source methods, significantly enhancing efficiency and reducing labor costs. As illustrated in Fig.~\ref{fig:app} (a) and (b), high-quality rendered videos effectively capture object details within the scenes and exhibit fewer aliasing artifacts. In contrast, method $A_1$ suffers from  noticeable spatial aliasing and temporal instability, leading to less satisfactory outcomes. In this application, although method \(A_4\) delivers superior results, its computational costs exceed our budget. The performance of \(O_3\) is comparable to that of \(A_3\), but it is achieved at a lower cost and within our budget. Using the proposed NR-VQA metric, we can rapidly and automatically validate the performance of the developed method.

\begin{table}
\caption{Quality assessment of various supersampling methods, presenting video quality scores (rescaled using the labeled OA-MOS) from our NR-VQA metric and human-labeled OA-MOS across two scenes.}
\centering
\centering\renewcommand{\arraystretch}{1.2} 
\footnotesize
\begin{tabular}{>{\centering\arraybackslash}p{1.4cm}|cc|cc} 
\hline\hline
\multirow{2}{*}{Methods} & \multicolumn{2}{c|}{Scene 1} & \multicolumn{2}{c}{Scene 2}\\
        & Ours $\uparrow$  & OA-MOS $\uparrow$  & Ours $\uparrow$  & OA-MOS $\uparrow$ \\ \hline 
$A_1$   & 1.72	& 1.65	& 1.89	& 1.82 \\
$A_2$   & 1.88	& 1.79	& 2.14	& 1.89 \\
$A_3$   & 2.89	& 2.85	& 2.92	& 3.18 \\
$A_4$   & \textbf{3.12}	& \textbf{3.71}	& \textbf{3.45}	& \textbf{3.41} \\ \hline
$O_1$   & 2.52	& 2.33	& 2.54	& 2.51 \\
$O_2$   & 2.71	& 2.52	& 2.74	& 2.79 \\
$O_3$   & \underline{2.99}	& \underline{3.02}	& \underline{3.01}	& \underline{3.19} \\
\hline\hline
\end{tabular}

\label{tab:app1}
\end{table}

\begin{table}
\caption{Quality assessment of frame generation strategies, showing scores from our NR-VQA metric (rescaled) and human-labeled OA-MOS.}
\centering
\centering\renewcommand{\arraystretch}{1.2} 
\footnotesize
\begin{tabular}{>{\centering\arraybackslash}p{1.6cm}|cc|cc} 
\hline\hline
\multirow{2}{*}{Strategies} & \multicolumn{2}{c|}{Scene 3} & \multicolumn{2}{c}{Scene 4}\\
        & Ours $\uparrow$  & OA-MOS $\uparrow$  & Ours $\uparrow$  & OA-MOS $\uparrow$ \\ \hline 
A-R      & 3.92  & 4.15  & 3.70  & 3.94\\ \hline 
1-I      & 3.63  & 3.71  & 3.41  & 3.48\\
1-E      & 3.46  & 3.44  & 3.23  & 3.07\\ \hline 
2-I      & 3.64  & 3.65  & 3.28  & 3.35\\
1-E/1-I  & 3.31  & 3.43  & 3.15  & 3.12\\
1-I/1-E  & 3.08  & 2.91  & 2.96  & 2.84\\
2-E      & 3.05  & 2.82  & 2.87  & 2.81\\
\hline\hline
\end{tabular}

\label{tab:app2}
\end{table}

\subsection{Evaluating Frame Generation Strategies}
Frame generation techniques (interpolation and extrapolation) are used in real-time rendering applications to enhance frame rates. Mob-FGSR~\cite{yang2024frame} introduces a method that can generate frames at arbitrary times, allowing for multiple interpolated and extrapolated frames. A significant challenge in deploying this algorithm is determining the optimal combination of interpolation and extrapolation to achieve the best video quality. Full-reference metrics are unsuitable since the estimated motion of generated frames typically do not align with reference images. Although user ratings provide a feasible method for quality assessment, they are labor-intensive and prone to inconsistency; as ratings from an individual can vary significantly from one day to the next. Thus, quantitative video quality assessments are crucial for efficient project development.

\begin{figure}
\begin{center}
\includegraphics[width=0.9\linewidth]{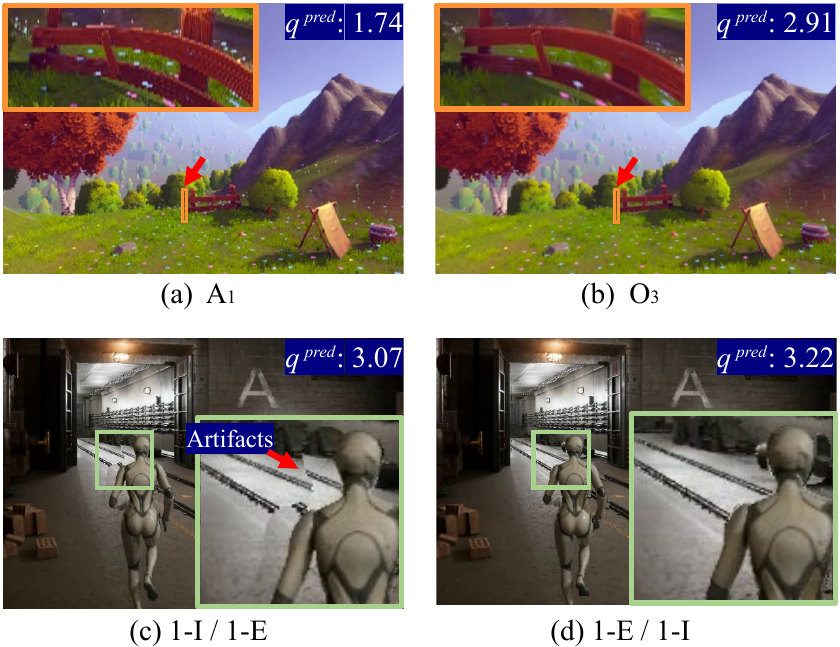}
\end{center}
\caption{Examples of the applications. (a) and (b) show the supersampling results, temporal profiles (orange boxes), and predicted video quality scores \( q^{{pred}} \). (c) and (d) illustrate the frame generation results and the predictions \( q^{{pred}} \), with artifacts shown in  light green boxes.}
\label{fig:app}
\end{figure}

In this study, we utilize our NR-VQA metric to quantitatively assess various frame generation strategies. We collect five 8-second videos at 60 FPS / 1080p for each of two 3D game scenes using different strategies: all rendered frames (``A-R''), one interpolated frame (``1-I''), one extrapolated frame (``1-E''), two interpolated frames (``2-I''), two extrapolated frames (``2-E''), interpolation followed by extrapolation (``1-I/1-E''), and extrapolation followed by interpolation (``1-E/1-I''). For each strategy, we gather videos that are closely similar in content for each scene and employ annotators to provide reference OA-MOS labels. The videos are then assessed for quality using our NR-VQA metric. The results, shown in Tab.~\ref{tab:app2}, depict the video quality for each strategy. Across both scenes, ``A-R'' yields the highest quality scores, followed by ``1-I'' and ``1-E'', with ``1-I'' outperforming ``1-E'' due to having additional frame references. For strategies involving two frame generations, ``2-I'' exhibits the best performance, followed by ``1-E/1-I'' and then ``1-I/1-E'' (see Fig. \ref{fig:app} (c) and (d)), with ``2-E'' showing the lowest quality. These results align with our expectations. Given the latency issues associated with interpolation, we recommend strategies ``1-E'' or ``1-E/1-I'' for latency-sensitive games, and ``1-I'' or ``2-I'' for games where delays are less sensitive.
 % which is calibrated on the ReVQ-2k (1080p) dataset

\section{Limitations}
Our NR-VQA dataset and metric for assessing rendered video quality have several limitations. First, the TS-MOS in our dataset primarily captures temporal instability but does not evaluate the fluidity or naturalness of object movements within the video. As a result, issues related to motion smoothness are not considered, which could be addressed in future work. 
Second, the NR-VQA metric shows varying sensitivity to different scene contents and motion velocities, potentially leading to biased scores in certain cases. To ensure a fair comparison, it is crucial that the scene contents and motion conditions of the compared videos are as consistent as possible. 
Third, stream (b) occasionally leads to less optimal outcomes. As seen in Fig.~\ref{fig:res1} panel (i), in cases where videos display good temporal stability but exhibit severe blurring, our method slightly overestimates the overall quality. To address this, future work could more precisely analyze the impact of temporal stability to enhance the accuracy of the metric.
Finally, due to the use of motion estimation, the model’s runtime is increased, typically taking over ten seconds to evaluate the quality of one video. This delay, however, remains acceptable for most practical applications.

\section{Conclusions}
Accurate video quality evaluations are essential for numerous rendering applications, such as pipeline optimization and parameter selection. To address the challenges of assessing the quality of rendered videos that cannot be perfectly aligned and lack reference videos, we introduce a large rendering-oriented VQA dataset along with a novel NR-VQA metric specifically designed for rendered content. The dataset, termed ReVQ-2k, comprises 2,000 videos featuring a variety of 3D scenes and rendering settings, each annotated with overall quality labels (OA-MOS) and temporal stability labels (TS-MOS). Our NR-VQA metric provides a comprehensive evaluation of rendered videos by analyzing both overall image quality and temporal stability. Experiments on the ReVQ-2k dataset confirm the superior accuracy of our metric, consistently outperforming existing SOTA methods. Furthermore, we demonstrate the practical utility of our NR-VQA metric in two real-world applications, showing its capacity to reduce manual labor and accelerate the development of rendering algorithms. Our work establishes a robust benchmark and provides a baseline method for NR-VQA of rendered videos, offering valuable insights for related applications and future research.

\ifCLASSOPTIONcaptionsoff
  \newpage
\fi

{
	\small
	\bibliographystyle{IEEEtran}
	\bibliography{ref}
}

\end{document}